\documentclass{article}
\usepackage{spconf,amsmath,hyperref,amssymb,array,booktabs}
\usepackage{enumitem}
\usepackage{booktabs}
\usepackage{tabularx}
\usepackage{multirow}
\usepackage{graphicx}
\usepackage[linesnumbered,ruled,vlined]{algorithm2e}
\usepackage{listings}
\usepackage{makecell}
\usepackage[backend=biber,style=ieee,
maxbibnames=6,
minbibnames=1,
]{biblatex}
\defbibheading{bibliography}[\refname]{}
\AtBeginBibliography{\setlength{\emergencystretch}{3em}}

\usepackage{tikz} 
\usetikzlibrary{shapes.geometric, arrows, positioning, calc} 
\tikzstyle{data} = [
    rectangle, rounded corners,
    minimum width=2cm, minimum height=1cm,
    text centered, align=center,
    draw=black, fill=gray!20
]

\tikzstyle{signal_encoder} = [
    rectangle,
    minimum width=1.6cm, minimum height=1cm,
    text centered, align=center,
    draw=black, fill=blue!15
]
\tikzstyle{text_encoder} = [
    rectangle,
    minimum width=2.cm, minimum height=1cm,
    text centered, align=center,
    draw=black, fill=blue!15
]

\tikzstyle{attention} = [
    rectangle,
    minimum width=1.5cm, minimum height=1cm,
    text centered, align=center,
    draw=black, fill=green!20
]

\tikzstyle{process} = [
    rectangle,
    minimum width=2cm, minimum height=1cm,
    text centered, align=center,
    draw=black, fill=orange!20
]

\tikzstyle{arrow} = [
    thick, ->, >=stealth, draw=black
]

\newlength{\jsonleftcol}
\newcommand{\jleft}[1]{\text{\makebox[\jsonleftcol][l]{#1}}}

\title{DiffNator: Generating Structured Explanations of Time-Series Differences}
\name{Kota Dohi, Tomoya Nishida, Harsh Purohit, Takashi Endo, Yohei Kawaguchi}
\address{Research and Development Group, Hitachi, Ltd.}

\begin{document}
\ninept
%
\maketitle
\begin{abstract}
In many IoT applications, the central interest lies not in individual sensor signals but in their differences, yet interpreting such differences requires expert knowledge. 
We propose \textit{DiffNator}, a framework for structured explanations of differences between two time series. We first design a JSON schema that captures the essential properties of such differences. 
Using the Time-series Observations of Real-world IoT (TORI) dataset, we generate paired sequences and train a model that combine a time-series encoder with a frozen LLM to output JSON-formatted explanations. 
Experimental results show that DiffNator generates accurate difference explanations and substantially outperforms both a visual question answering (VQA) baseline and a retrieval method using a pre-trained time-series encoder.
\end{abstract}
\section{Introduction}
\label{sec:intro}
The spread of the Internet of Things (IoT) has enabled large-scale collection of sensor data from industrial machinery.
However, analyzing and interpreting such data often requires domain expertise, posing challenges for non-experts \cite{aksu2024xforecast, guo2024vivar}. To address this issue, prior studies have proposed methods that automatically generate natural-language explanations from sensor data \cite{reiter2007architecture, kobayashi2013probabilistic, murakami2017learning, hamazono-etal-2020-market, dohi2025TACO}.

Existing approaches to explanation generation from sensor data have primarily focused on describing individual time series, either in isolation or as multiple independent signals. In many practical scenarios, however, the central interest lies in the differences between signals. Typical examples include comparisons across heterogeneous sensors or between normal and abnormal operation. Such cases highlight the limitations of conventional methods, which are not designed to explicitly capture and explain differences.

In this work, we propose DiffNator, a framework for generating natural-language explanations of differences between two time series. We focus on two types of differences: Type 1, where a phenomenon appears in only one of the two series, and Type 2, where it appears in both series but with different intensity. To reproduce these differences, we construct paired data from a common time series by applying functions that represent specific phenomena: for Type 1, the function is added to only one of the paired series, while for Type 2, the function is applied to both series with a varied parameter. We design a unified JSON schema that captures the essential information for both types. Building on prior work in time-series question answering \cite{chow2024timeseriesreasoning}, 
we apply a model that integrates a time-series encoder to produce embeddings, which are fused with textual prompts and fed into a large language model (LLM). The model is trained to generate explanations in accordance with the proposed schema, thereby enabling systematic evaluation and seamless integration with downstream systems.
To validate the proposed framework, we introduce the TORI (Time-series Observations of Real-world IoT) dataset. Paired time series derived from TORI are used to train the model to generate JSON-formatted difference explanations, which are subsequently evaluated.



\begin{table}[t]
\centering
\caption{Examples of reference--target time-series pairs with structured difference explanations. The three columns show (i) time-series plots (blue: reference, red: target), (ii) ground-truth lists of JSON formats, and (iii) generated outputs from our model (using the Informer encoder with query attention on difference).}
\setlength{\tabcolsep}{6pt}
\begin{tabular}{m{3.1cm}@{\hspace{0pt}}m{2.5cm}@{\hspace{2pt}}m{2.5cm}}
\toprule
\textbf{Time-series data} & \textbf{Ground truth} & \textbf{Generated} \\
\midrule

\hspace*{-2mm}\includegraphics[width=\linewidth]{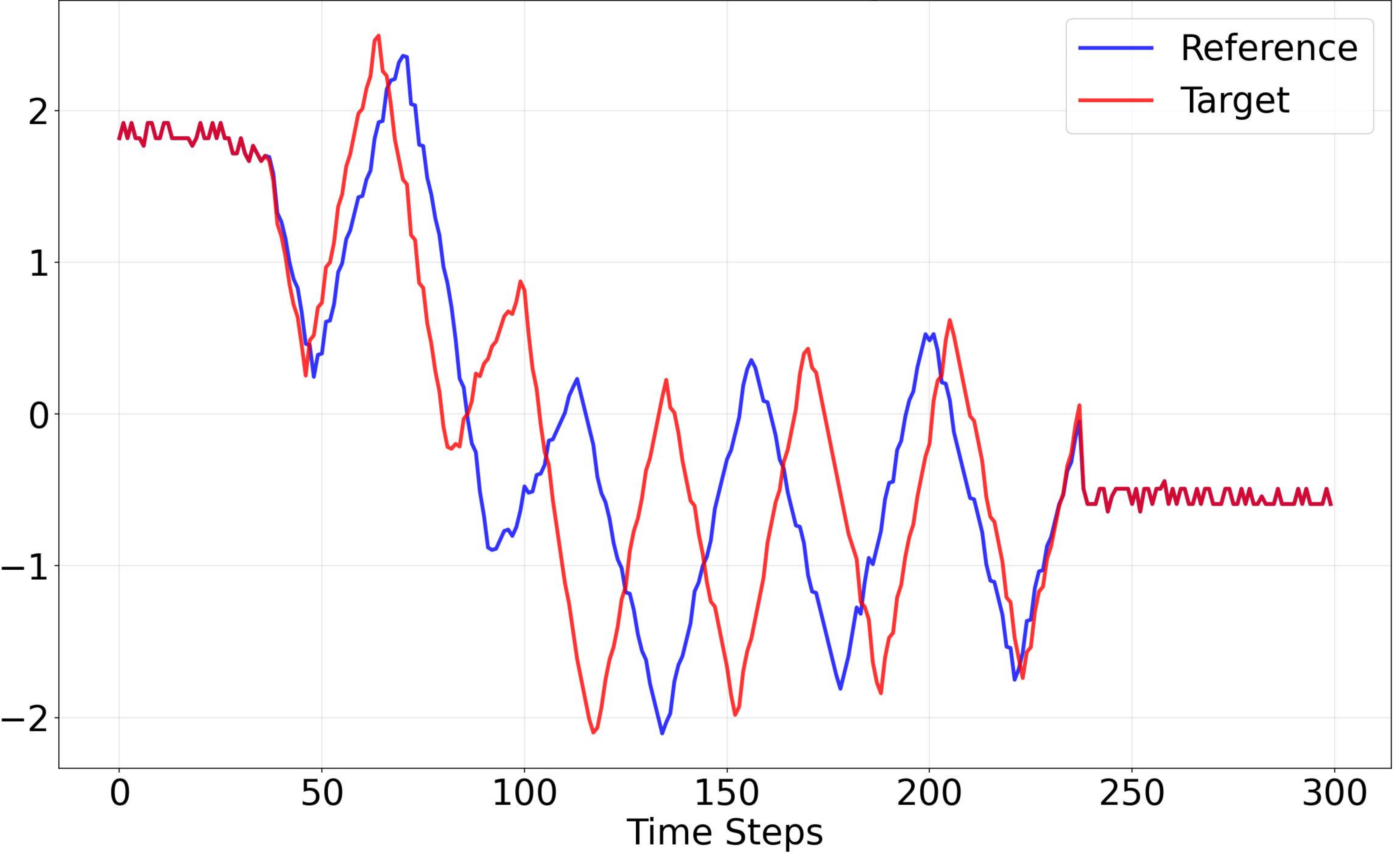} &
\begin{minipage}{\linewidth}
\vspace{-7mm}
\begin{lstlisting}
[{
 type:TYPE2,
 func:TRIANGLE_WAVE,
 start:36,
 end:237,
 param:FREQUENCY,
 magnitude:LARGER}]
\end{lstlisting}
\end{minipage} &
\begin{minipage}{\linewidth}
\vspace{-5mm}
\begin{lstlisting}
[{
 type:TYPE2,
 func:TRIANGLE_WAVE,
 start:47,
 end:237,
 param:FREQUENCY,
 magnitude:LARGER}]
\end{lstlisting}
\end{minipage} \\
\midrule

\vspace{2mm}
\hspace*{-2mm}\includegraphics[width=\linewidth]{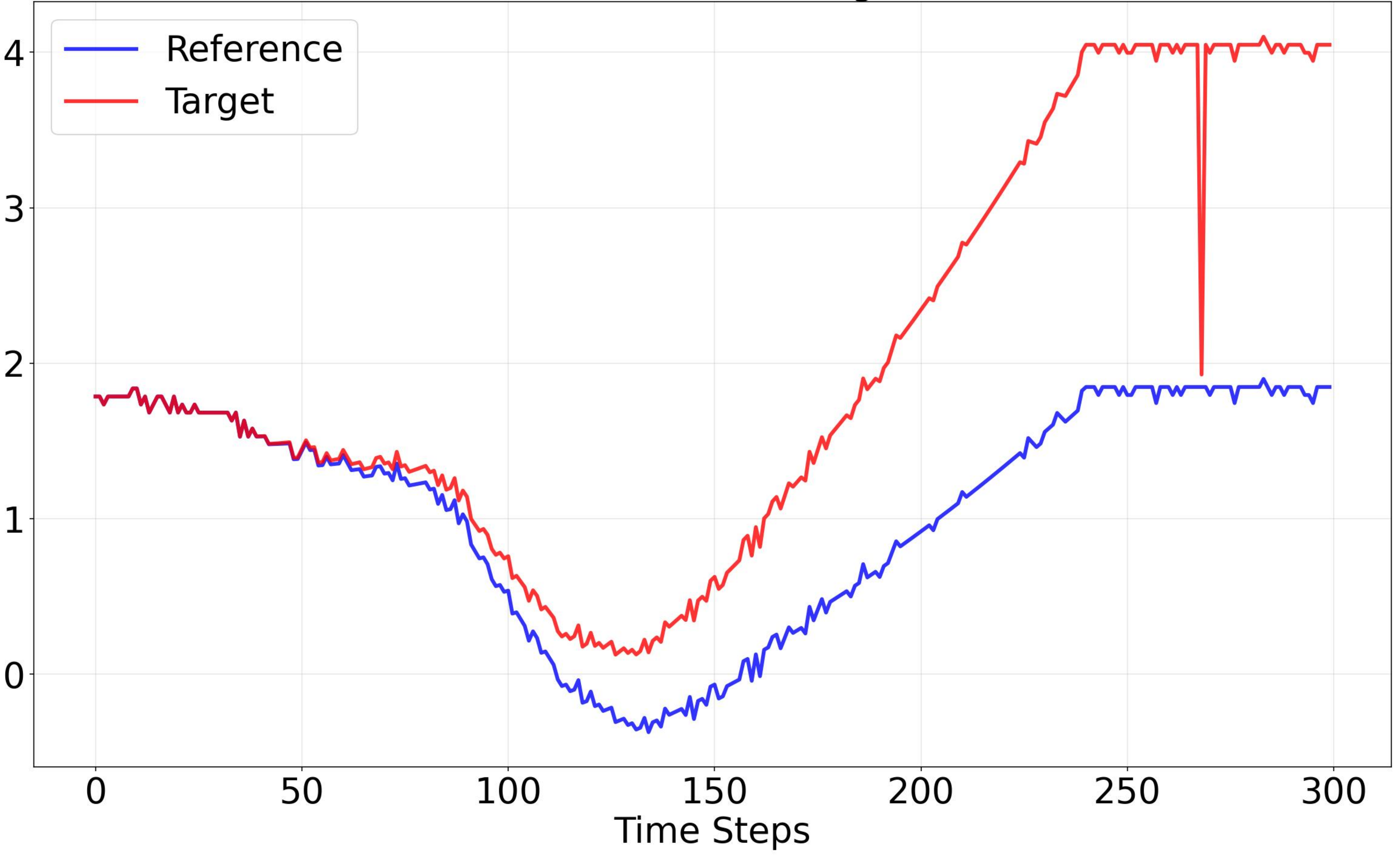} &
\begin{minipage}{\linewidth}
\vspace{-5mm}
\begin{lstlisting}
[{
 type:TYPE2,
 func:QUADRATIC_INCREASE,
 start:35,
 end:240,
 param:AMPLITUDE,
 magnitude:LARGER},
 {
 type:TYPE1,
 func:DROP,
 start:268,
 end:268,
 presence:PRESENT}]
\end{lstlisting}
\end{minipage} &
\begin{minipage}{\linewidth}
\vspace{-5mm}
\begin{lstlisting}
[{
 type:TYPE2,
 func:QUADRATIC_INCREASE,
 start:40,
 end:240,
 param:AMPLITUDE,
 magnitude:LARGER},
 {
 type:TYPE1,
 func:DROP,
 start:268,
 end:268,
 presence:PRESENT}]
\end{lstlisting}
\end{minipage} \\
\bottomrule
\end{tabular}
\label{tab:json_examples}
\end{table}

\vspace{-4mm}
\section{Relation to prior work}
\label{sec:format}

Early data-to-text studies trained task-specific models to generate natural-language descriptions of individual time series, often in a domain-dependent manner \cite{reiter2007architecture, kobayashi2013probabilistic, murakami2017learning, hamazono-etal-2020-market}. 
To overcome this limitation, our prior work proposed a domain-independent framework \cite{dohi2025TACO} that leverages pre-trained text encoders and decoders, with learning concentrated on a time-series encoder to describe diverse characteristics.
The advent of time-series foundation models has enabled general-purpose forecasting, anomaly detection, and representation learning \cite{ansari2024chronos,das2023timesfm,goswami2024moment,woo2024moirai,garza2023timegpt, ekambaram2025tspulse}. 
Building on this trend, time-series question answering (TS-QA) models have emerged, integrating LLMs with specialized time-series encoders to explain time-series characteristics in response to diverse prompts \cite{chow2024timeseriesreasoning, Wang2025ITFormer, Xie2025ChatTS, Kong2025TimeMQA}. 
Another related line of research is chart-to-text generation, which develops captioning methods for chart images, including those containing multiple time series \cite{Obeid2020Chart2Text, Mahinpei2022LineCap, Kantharaj2022Chart2Text, Tang2023VisText}. 
However, these lines of prior work have mainly focused on describing individual series or making simple value-level comparisons between multiple time series, without providing systematic explanations of inter-series differences.

In contrast, our work explicitly targets inter-series differences and develops a framework for their systematic explanation.


\section{Problem Statement}
\label{sec:problem}

\noindent\textbf{Setting.}
Let $x_{\mathit{ref}}, x_{\mathit{tgt}} \in \mathbb{R}^{T}$ be a pair of univariate time series of equal length $T$, called the \emph{reference data} and \emph{target data}, respectively.
Define the pointwise difference
\[
\Delta(t) \;=\; x_{\mathit{tgt}}(t) - x_{\mathit{ref}}(t), \quad t=1,\dots,T .
\]
We interpret $\Delta(t)$ as the superposition of $K$ physical phenomena. 
Each phenomenon, called an \emph{elementary difference}, is modeled by a component function chosen from a pre-defined library 
$\{\, g_\ell(\,\cdot\,;\boldsymbol{\theta}) \mid \ell \in \mathcal{L} \,\}$.
The parameter vector $\boldsymbol{\theta}$ 
is taken from an admissible set $\Theta_\ell \subset \mathbb{R}^{d_\ell}$, 
whose dimension and semantics depend on $\ell$.
Each elementary difference $k$ is characterized by a component function $g_{\ell_k}(t;\boldsymbol{\theta})$, 
parameter vectors $\boldsymbol{\theta}_{\mathit{ref}}^{(k)}, \boldsymbol{\theta}_{\mathit{tgt}}^{(k)} \in \Theta_{\ell_k}$, 
activity indicators $\delta_{\mathit{ref}}^{(k)}, \delta_{\mathit{tgt}}^{(k)} \in \{0,1\}$, 
and an active interval
\[
I_k \;=\; \{\, t \in \{1,\dots,T\} \mid \tau_s^{(k)} \le t \le \tau_e^{(k)} \,\}.
\]
Intervals $I_k$ are not required to be disjoint; multiple phenomena may overlap in time.
The pointwise difference can then be expressed as
\begin{equation}
\Delta(t)
\;=\;
\sum_{k=1}^K \mathbf{1}[t \in I_k]\Big(\,\delta_{\mathit{tgt}}^{(k)}\, g_{\ell_k}(t;\boldsymbol{\theta}_{\mathit{tgt}}^{(k)})
\;-\;\delta_{\mathit{ref}}^{(k)}\, g_{\ell_k}(t;\boldsymbol{\theta}_{\mathit{ref}}^{(k)})\Big),
\label{eq:diff-multi}
\end{equation}
where $\mathbf{1}[\cdot]$ is the indicator function; hence $\Delta(t)=0$ for $t \notin \bigcup_{k=1}^K I_k$.
To exclude degenerate cases, we assume for every $k$ that $\delta_{\mathit{ref}}^{(k)} + \delta_{\mathit{tgt}}^{(k)} \ge 1$.

\noindent\textbf{Structured difference explanation.}
Given $(x_{\mathit{ref}}, x_{\mathit{tgt}})$, the output is defined as a list of JSON objects, one for each elementary difference.
Each object is defined by the following fields, which capture how the target data differs from the reference data:

\noindent\textbf{\texttt{type:}} the difference type, either Type~1 or Type~2.
\begin{itemize}
  \item \emph{Type~1:} 
  $\delta_{\mathit{ref}}^{(k)} \neq \delta_{\mathit{tgt}}^{(k)}$, i.e., the phenomenon is active in only one series.
  \item \emph{Type~2:} 
  $\delta_{\mathit{ref}}^{(k)} = \delta_{\mathit{tgt}}^{(k)} = 1$ and the parameter vectors differ in at most one entry; 
  that is, there exists $p$ such that
  \[
  \boldsymbol{\theta}_{\mathit{ref}}^{(k)}[p] \neq \boldsymbol{\theta}_{\mathit{tgt}}^{(k)}[p],
  \quad
  \boldsymbol{\theta}_{\mathit{ref}}^{(k)}[p'] = \boldsymbol{\theta}_{\mathit{tgt}}^{(k)}[p'] \ \text{for all } p' \ne p .
  \]
\end{itemize}
\noindent\textbf{\texttt{func:}} the name of the component function $g_{\ell_k}$. 

\noindent\textbf{\texttt{start, end:}} the start and end indices $(\tau_s^{(k)}, \tau_e^{(k)})$ of $I_k$.

\noindent\textbf{\texttt{presence:}} whether the component function appears in $x_{\mathit{tgt}}$.
\[
\begin{cases}
  \jleft{\texttt{PRESENT}} & \text{if }\delta_{\mathit{tgt}}^{(k)}=1 \land \text{Type~1},\\
  \jleft{\texttt{ABSENT}}  & \text{if }\delta_{\mathit{tgt}}^{(k)}=0 \land \text{Type~1},\\
  \jleft{\texttt{null}}      & \text{otherwise}.
\end{cases}
\]
\noindent\textbf{\texttt{param:}} the name of the parameter that differs between $x_{\mathit{ref}}$ and $x_{\mathit{tgt}}$.
\[
\begin{cases}
  \jleft{\text{name of parameter $p$}} &
  \text{if } \boldsymbol{\theta}_{\mathit{ref}}^{(k)}[p] \neq \boldsymbol{\theta}_{\mathit{tgt}}^{(k)}[p] \land \text{Type~2},\\
  \jleft{\texttt{null}} & \text{otherwise}.
\end{cases}
\]
\noindent\textbf{\texttt{magnitude:}} the relative magnitude of the differing parameter $p$.
\[
\begin{cases}
  \jleft{\texttt{LARGER}}  &
  \text{if } \bigl|\boldsymbol{\theta}_{\mathit{tgt}}^{(k)}[p]\bigr| > \bigl|\boldsymbol{\theta}_{\mathit{ref}}^{(k)}[p]\bigr| \land \text{Type~2},\\
  \jleft{\texttt{SMALLER}} &
  \text{if } \bigl|\boldsymbol{\theta}_{\mathit{tgt}}^{(k)}[p]\bigr| < \bigl|\boldsymbol{\theta}_{\mathit{ref}}^{(k)}[p]\bigr| \land \text{Type~2},\\
  \jleft{\texttt{null}} & \text{otherwise}.
\end{cases}
\]

\begin{table}[t]
  \caption{Component functions by category (shown in Capitalized style for readability; actual JSON use uppercase).}
  \centering
  \footnotesize
  \renewcommand{\arraystretch}{1.05}
  \begin{tabular}{|>{\raggedright\arraybackslash}p{0.15\linewidth}|>{\raggedright\arraybackslash}p{0.72\linewidth}|}
    \hline
    \textbf{Category} & \textbf{Component functions} \\
    \hline
    Trend &
    Linear\_increase, Linear\_decrease, Quadratic\_increase, Quadratic\_decrease, 
    Cubic\_increase, Cubic\_decrease, Exponential\_growth, Inverted\_exponential\_growth, 
    Exponential\_decay, Inverted\_exponential\_decay, Log\_increase, Log\_decrease, 
    Sigmoid, Inverted\_sigmoid, Gaussian, Inverted\_gaussian \\
    \hline
    Periodic &
    Sinusoidal, Sawtooth, Square\_wave, Triangle\_wave \\
    \hline
    Fluctuation &
    Gaussian\_noise, Laplace\_noise \\
    \hline
    Event &
    Spike, Drop, Positive\_step, Negative\_step, Positive\_pulse, Negative\_pulse \\
    \hline
  \end{tabular}
  \label{tab:components}
\end{table}

\section{Data Generation Protocol}
\label{sec:data_gen_protocol}

\noindent\textbf{Source corpus.}
To construct reference–target pairs, we built the TORI (Time-series Observations of Real-world IoT) dataset using 1,690,485 real-world time-series samples collected from sensors. From this dataset, we randomly selected 1,000 samples for the validation set and another 1,000 for the test set. All remaining samples were used for the training set.

\noindent\textbf{Component function taxonomy and bounds.}
We prepared 27 component functions, categorized into four types—Trend, Periodic, Fluctuation, and Event; 
Table~\ref{tab:components} lists all functions by category.
This taxonomy was largely inspired by the synthetic time-series dataset SUSHI~\cite{kawagu_sushi}.
Unlike SUSHI, however, we further divided the original fluctuation category into two: 
\emph{Fluctuation}, which includes components such as noise that persist over relatively long intervals, and 
\emph{Event}, which covers phenomena such as steps or spikes that occur only within short intervals on the time axis.
Each component function $g_{\ell}$ is associated with duration bounds 
$L_{\min}^{\ell}$ and $L_{\max}^{\ell}$, which specify the admissible 
length of any active interval. 
Specific values for these bounds are available in the project page \cite{dohi_diffnator}.


\noindent\textbf{Pair generation protocol.}  
We first obtain a raw time series from the TORI sensor corpus. Its length is adjusted to $T$ by resampling if shorter and truncation if longer. After applying $z$-normalization, the resulting time series data $x$ is used as the baseline for subsequent modifications. Based on this baseline, we generate a pair of time series $(x_{\mathit{ref}}, x_{\mathit{tgt}})$ by introducing a random number of elementary differences, sampled uniformly from the predefined range $[K_{\min}, K_{\max}]$. 
For each $k \in \{1,\dots,K\}$, we first sample a category $c_k$ uniformly from $\{\text{Trend}, \text{Periodic}, \text{Fluctuation}, \text{Event}\}$ and then randomly select a component function $g_{\ell_k}$ from the set associated with $c_k$. The chosen function is instantiated with base parameters $\boldsymbol{\theta}_{\mathrm{base}}^{(k)}$ sampled independently from predefined ranges, and the interval $I_k$ is determined by randomly sampling both its length and its start index.  

Next, we sample a difference type $\kappa^{(k)}$ uniformly from \{Type~1, Type~2\}.  
If $\kappa^{(k)}=\text{Type 1}$, the function is added to exactly one of $\{x_{\mathit{ref}}, x_{\mathit{tgt}}\}$, with the series chosen uniformly at random.  
If $\kappa^{(k)}=\text{Type 2}$, we first uniformly choose a parameter index $p^{(k)} \in \{1,\dots,d_{\ell_k}\}$. A component-specific predefined rule specifies whether the modification is performed by adding an offset value $\eta > 0$ or by multiplying a ratio $\rho > 1$. The value is then sampled uniformly from its predefined range, and a modified parameter vector $\boldsymbol{\theta}_{\mathrm{mod}}^{(k)}$ is obtained by altering only the $p^{(k)}$-th entry of $\boldsymbol{\theta}_{\mathrm{base}}^{(k)}$. The function is then added to both series, with one using the base and the other the modified parameters, assigned uniformly at random.
The predefined ranges for sampling base parameters, the offset and ratio values are provided on the project page \cite{dohi_diffnator}.

\section{DiffNator framework}

\begin{figure}[t]
\centering
\tikzset{font=\fontsize{14}{14}\selectfont} 
\resizebox{1.0\linewidth}{!}{
\begin{tikzpicture}[
  font=\small,
  node distance = 9mm and 7mm,
  >=latex,
  tsdata/.style   = {fill=blue!10},      
  tsmod/.style    = {fill=blue!20},      
  textdata/.style = {fill=green!10},     
  textmod/.style  = {fill=green!20},     
  fusionfill/.style = {fill=gray!20, draw=none},  
  modelfill/.style  = {fill=violet!25},  
  box/.style   = {draw, rounded corners, minimum height=7mm, inner xsep=2mm},
  data/.style  = {box, draw=none},
  module/.style= {box},
  output/.style= {box, draw=none},
  trainable/.style = {line width=1.0pt},        
  frozen/.style    = {line width=1.0pt, dashed} 
]

\node[data, tsdata] (ref) {\shortstack{Reference \\ data}};
\node[data, tsdata, below=of ref] (tgt) {\shortstack{Target \\ data}};

\node[module, tsmod, trainable,  right=3mm of ref] (tsenc1) {\shortstack{TS \\ encoder}};
\node[module, tsmod, trainable,  right=5mm of tgt] (tsenc2) {\shortstack{TS \\ encoder}};

\node[module, tsmod, trainable, right=12mm of $(tsenc1)!0.5!(tsenc2)$] (merge) {Merge};
\node[module, tsmod, trainable, right=3mm of merge] (adapter) {Adapter};
\node[data, tsdata, trainable, right=3mm of adapter] (tsemb) {\shortstack{TS \\ embedding}};

\node[module, fusionfill, trainable, below=3mm of tsemb] (fusion) {Fusion};
\node[module, textmod, frozen,    right=3mm of fusion] (llm) {LLM};
\node[output, textdata, trainable,  right=3mm of llm] (json) {\shortstack{JSON \\ output}};

\node[data, textdata, trainable, below=3mm of fusion] (txtemb) {\shortstack{Text \\ embedding}};
\node[module, textmod, frozen, left of=txtemb, xshift=-1.2cm] (tok) {\shortstack{Tokenizer \\ + encoder}};
\node[data, textdata, trainable, left of=tok, xshift=-1.0cm] (prompt) {\shortstack{Prompt \\ text}};

\draw[->] (ref) -- (tsenc1);
\draw[->] (tgt) -- (tsenc2);
\draw[->] (tsenc1) -- (merge);
\draw[->] (tsenc2) -- (merge);
\draw[->] (merge) -- (adapter);
\draw[->] (adapter) -- (tsemb);

\draw[->] (prompt) -- (tok);
\draw[->] (tok) -- (txtemb);

\draw[->] (tsemb) -- (fusion);
\draw[->] (txtemb) -- (fusion);
\draw[->] (fusion) -- (llm);
\draw[->] (llm) -- (json);

\begin{scope}[shift={($(current bounding box.north east)+(-22mm,5mm)$)}]






  \draw[trainable] (0,-42mm) -- +(7mm,0);
  \node[anchor=west] at (9mm,-42mm) {Trainable};

  \draw[frozen] (0,-47mm) -- +(7mm,0);
  \node[anchor=west] at (9mm,-47mm) {Frozen};

\end{scope}
\end{tikzpicture}
}
\caption{Overview of the DiffNator framework. Reference and target time series data are individually encoded, merged, and projected via an adapter into LLM-compatible embeddings. The resulting embedding is fused with the prompt text embedding and fed into the LLM, which generates structured outputs in JSON format.}
\label{fig:model_architecture}
\end{figure}
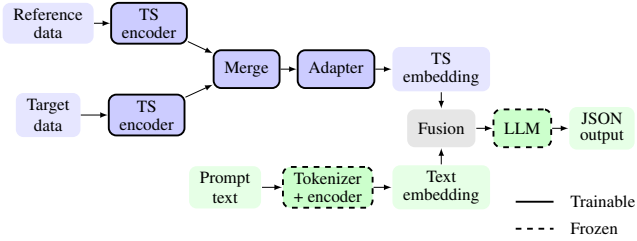

Figure~\ref{fig:model_architecture} illustrates the DiffNator framework, which integrates time-series encoders, difference-aware merging, transformation into embeddings interpretable by the LLM, and prompt-based conditioning.

\smallskip\noindent
(i) A time-series encoder (Informer~\cite{zhou2021informer} or Dilated~TCN~\cite{Lea2017TCN, bai2018empiricalevaluationgenericconvolutional}) maps the reference and target data into sequence of hidden representations 
$H_{\mathrm{ref}}, H_{\mathrm{tgt}} \in \mathbb{R}^{T\times D}$.

\smallskip\noindent
(ii) A merging module fuses the two sequences into a difference-aware representation. We implement two schemes:

\begin{enumerate}
 \item \textit{Mean difference pooling.}
 Let $D_t = H_{\mathrm{tgt}}(t) - H_{\mathrm{ref}}(t)$. The final representation is the temporal mean
 \[
   z = \frac{1}{T}\sum_{t=1}^{T} D_t \in \mathbb{R}^{D}.
 \]

 \item \textit{Query attention on difference.}
 Let $Q \in \mathbb{R}^{J\times D}$ be $J$ learnable queries that attend to $\{D_t\}_{t=1}^{T}$. We compute
 \[
   \alpha_{j,t} = \operatorname{softmax}_{t}\!\bigl(Q_j D_t^{\top}\bigr),\;
   z_j = \sum_{t=1}^{T} \alpha_{j,t} D_t,\ \ j=1,\dots,J,
 \]
 and stack the query outputs to obtain $Z = [z_1;\dots;z_J] \in \mathbb{R}^{J\times D}.$
 This highlights localized differences.
\end{enumerate}


\smallskip\noindent
(iii) An adapter with a 2-layer MLP (GELU + LayerNorm) converts the representation into an LLM-compatible embedding.

\smallskip\noindent
(iv) A fusion module inserts the projected embedding into the prompt-text embeddings and forwards the combined sequence to the LLM to produce JSON outputs. 
The prompt text has three parts: \emph{inst\_part1}, \emph{inst\_part2} and \emph{question}.
\emph{inst\_part1} is a part of the instruction that introduces the task and indicates the input that follows:
\texttt{"Task: You are a time-series difference explainer. Input:"}.
\emph{inst\_part2} defines the output schema by listing the required JSON fields (e.g., \texttt{type}, \texttt{func}, \texttt{start}, \texttt{end}) and requires the output to be strictly valid JSON. 
Finally, the \emph{question} is a short request to explain the differences between the two time-series data.
In constructing the LLM input, the time-series embedding is inserted between the embeddings of \emph{inst\_part1} and \emph{inst\_part2}, 
so that it is interpreted as part of the input.

The time-series encoder, the merging module (for query attention), 
and the adapter are all trainable. 
The LLM parameters are kept frozen, and the model is optimized end-to-end with respect to the trainable modules only.
The training objective is the causal language modeling loss, computed only on the answer tokens.

\begin{table*}[t]
\centering
\caption{Performance of compared approaches under the $K_{\max}{=}1$ setting. 
All values are reported as percentages. 
We evaluate our DiffNator framework with two encoders (Informer, Dilated~TCN) and two merging methods 
(mean difference pooling, denoted as \textbf{mean}; query attention on difference, denoted as \textbf{attn}), 
and compare against two baselines: a visual question answering approach using GPT-4o (VQA) and a retrieval-based method using 
TSPulse embeddings (GT-search). 
Results for DiffNator are averaged over three runs.}
\label{tab:results_k1}
\begin{tabular}{l@{\hskip 8pt}c@{\hskip 10pt}c@{\hskip 8pt}c@{\hskip 5pt}c@{\hskip 5pt}c@{\hskip 5pt}c@{\hskip 8pt}c|c@{\hskip 8pt}c@{\hskip 5pt}c@{\hskip 5pt}c}
\toprule
\multirow{2}{*}{\textbf{Approach}} 
& \multicolumn{5}{c}{\textbf{Field accuracies (overall)}} 
& \multirow{2}{*}{\makecell{\textbf{IoU} \\ \textbf{(overall)} }} 
& \multirow{2}{*}{\makecell{\textbf{Match acc.} \\ \textbf{(overall)}}} 
& \multicolumn{4}{c}{\textbf{Match acc. (by category)}} \\
\cmidrule(lr){2-6} \cmidrule(lr){9-12}
 & Type & Func & Presence & Param & Magnitude &  &  & Trend & Periodic & Fluctuation & Event \\
\midrule
GPT-4o (VQA)       & 60.4 & 30.6 & 62.3 &  8.4 & 26.3 & 36.8 &  3.9 & 0.9 & 14.6 & 5.1 & 0.0 \\
TSPulse (GT-search)       & 89.9 & 60.8 & 89.2 &  76.7 & 72.2 & 60.2 &  25.5 & 32.7 & 36.8 & 10.5 & 23.7 \\
\midrule
Dilated~TCN + mean          & 99.3 & 94.1 & 98.7 & 98.2 & 99.3 & 89.0 & 83.3 & 76.0 & 96.6 & 95.9 & 64.6 \\
Dilated~TCN + attn          & 99.2 & 97.3 & 98.6 & \textbf{99.7} & \textbf{99.4} & \textbf{91.9} & 89.1 & \textbf{96.4} & 97.4 & 91.8 & \textbf{70.9} \\
Informer + mean             & 98.9 & 90.7 & 98.0 & 97.2 & 98.7 & 84.7 & 75.7 & 65.1 & 94.5 & 94.6 & 48.4 \\
Informer + attn             & \textbf{99.4} & \textbf{97.7} & \textbf{99.2} & 99.3 & 99.3 & 91.4 & \textbf{89.4} & 93.4 & \textbf{98.3} & \textbf{96.3} & 69.4 \\
\bottomrule
\end{tabular}
\end{table*}

\begin{table*}[t]
\centering
\caption{Performance of our DiffNator framework with Informer+attn under the $K_{\max}{=}2,3,4$ settings. 
All values are reported as percentages. Results are averaged over three runs.}
\label{tab:results_k2}
\begin{tabular}{l@{\hskip 8pt}c@{\hskip 10pt}c@{\hskip 8pt}c@{\hskip 5pt}c@{\hskip 5pt}c@{\hskip 5pt}c@{\hskip 8pt}c@{\hskip 8pt}c@{\hskip 8pt}c|c@{\hskip 8pt}c@{\hskip 5pt}c@{\hskip 5pt}c}
\toprule
\multirow{2}{*}{\textbf{$K_{\max}$}} 
& \multicolumn{5}{c}{\textbf{Field accuracies (overall)}} 
& \multirow{2}{*}{\makecell{\textbf{IoU} \\ \textbf{(overall)}}} 
& \multirow{2}{*}{\makecell{\textbf{Match acc.} \\ \textbf{(overall)}}} 
& \multirow{2}{*}{\textbf{OPR}} 
& \multirow{2}{*}{\textbf{UPR}} 
& \multicolumn{4}{c}{\textbf{Match acc. (by category)}} \\
\cmidrule(lr){2-6} \cmidrule(lr){11-14}
 & Type & Func & Presence & Param & Magnitude &  &  &  &  & Trend & Periodic & Fluctuation & Event \\
\midrule
2 & 97.9 & 91.7 & 98.1 & 96.0 & 97.4 & 87.3 & 78.4 &  3.0 &  0.7 & 68.0 & 94.8 & 93.1 & 59.6 \\
3 & 96.6 & 88.3 & 96.4 & 92.8 & 94.8 & 84.6 & 72.3 &  5.0 &  1.0 & 51.0 & 94.6 & 91.7 & 50.2 \\
4 & 95.6 & 83.5 & 94.5 & 90.9 & 92.0 & 81.4 & 66.4 &  5.6 &  1.7 & 36.8 & 92.0 & 91.6 & 42.6 \\
\bottomrule
\end{tabular}
\end{table*}

\section{Experiments}
\label{sec:experiments}

\subsection{Experimental Conditions}

\noindent\textbf{Data generation.}
We used the data generation protocol described in section \ref{sec:data_gen_protocol}.
Training data were synthesized on-the-fly: each epoch comprises $10{,}000$ freshly generated reference/target pairs.
Validation and test sets were comprised of fixed $1{,}000$ pairs each.

\medskip\noindent
\textbf{Model.}
The LLM backbone was fixed to \textit{Mistral-7B-Instruct-v0.1}\footnote{https://huggingface.co/mistralai/Mistral-7B-Instruct-v0.1}.
For the time-series encoder, we evaluated Informer and Dilated~TCN.
Informer was configured with three encoder layers, each consisting of attention and convolution blocks, with hidden size 512. Dilated~TCN was configured with eight convolutional layers with kernel size 3 and hidden size 512.
Difference-aware merging was performed either by mean difference pooling or query attention on difference with $J=6$ queries.
The resulting vector(s) were mapped to the LLM embedding dimension of $4,096$ by the adapter that has 2-layer MLP with GELU activation and LayerNorm.
We trained with the AdamW optimizer~\cite{loshchilov2019decoupled}, batch size $2$, and learning rate $1\times10^{-4}$ applied to all trainable modules.
A cosine annealing scheduler with warm restarts was used for a total of 500 epochs.

\medskip\noindent
\textbf{Compared approaches.}
We evaluated the DiffNator framework in four configurations, 
combining two encoders (Informer and Dilated~TCN) with two merging methods 
(mean difference pooling and query attention on difference). 
As baselines, we considered (i) a visual question answering (VQA) approach using GPT-4o \cite{openai2024gpt4ocard}, 
and (ii) a retrieval-based method using the pre-trained TSPulse \cite{ekambaram2025tspulse}.

For GPT-4o, we provided a single plot in which both the reference and target series were overlaid 
(as illustrated in Table~\ref{tab:json_examples}) and prompted it to generate JSON outputs, 
with temperature fixed to 0 for deterministic inference. 
For TSPulse, each time series was embedded into a 240-dimensional vector, 
and a difference embedding was obtained by subtracting the reference and target embeddings. 
We also generated a retrieval pool of 5 million time-series pairs on the fly from the training set 
and computed their difference embeddings. 
Each test embedding was matched to the most similar pool embedding by cosine similarity, 
and the ground-truth JSON object associated with the retrieved sample was returned as the output.

Evaluations were conducted under different settings of the maximum number of elementary differences, 
from $K_{\max}=1$ to $K_{\max}=4$, while keeping $K_{\min}=1$ fixed.

\subsection{Evaluation Metrics}
For each generated list of JSON objects, we aligned predicted and ground-truth elements before computing metrics. 
For $K_{\max}>1$, we greedily matched elements with the same component category (Trend, Periodic, Fluctuation, Event). If no such matches were found, we fell back to a position-based matching according to the order of elements in the JSON list. Remaining elements were treated as unmatched. Given the set of matched pairs, we computed the following metrics:

\begin{itemize}[leftmargin=1.5em]
  \item \textbf{Field-wise accuracies}: proportion of matched pairs for which each field was correctly generated. We evaluated \texttt{type} and \texttt{func} for all cases. We additionally evaluated \texttt{presence} for ground-truth elements with Type~1, and \texttt{param} and \texttt{magnitude} for those with Type~2.

  \item \textbf{Interval IoU (IoU)}: for each matched pair, we computed the intersection-over-union between the predicted interval $[\hat{\tau}_s,\hat{\tau}_e]$ and the ground-truth interval $[\tau_s,\tau_e]$:
  \[
    \text{IoU} = \frac{\max(0, \min(\hat{\tau}_e,\tau_e) - \max(\hat{\tau}_s,\tau_s))}
                      {\max(\hat{\tau}_e,\tau_e) - \min(\hat{\tau}_s,\tau_s)}.
  \]
  We then averaged IoU across all matched pairs and report it as a percentage for consistency with other metrics.

  \item \textbf{Match accuracy}: proportion of matched pairs for which all fields except \texttt{start/end} were correctly generated and the interval IoU was at least 0.8. We reported this metric overall and also separately according to the ground-truth component function category (Trend, Periodic, Fluctuation, Event).
\end{itemize}

In addition, when $K_{\max}>1$, we reported over- and under-prediction rates based
on unmatched elements after alignment. Specifically, the over-prediction
rate (OPR) and the under-prediction rate (UPR) were defined as
\[
\text{OPR} = |\text{unmatched}_{\mathrm{gen}}|/{|O|},
\quad
\text{UPR} = |\text{unmatched}_{\mathrm{gt}}|/|O|,
\]
where $|O|$ denotes the number of ground-truth elements in the test dataset,
and $|\text{unmatched}_{\mathrm{gen}}|$ and $|\text{unmatched}_{\mathrm{gt}}|$ are the numbers of unmatched predicted and ground-truth elements, respectively.

\subsection{Results}
All metrics, except for those of the VQA baseline, were averaged over three independent runs.

\noindent\textbf{Performance under $K_{\max}=1$.}
Table~\ref{tab:results_k1} compares two baselines, GPT-4o (VQA) and TSPulse (GT-search), with our four DiffNator configurations.
GPT-4o achieved 60.4\% accuracy on \texttt{type}, and 62.3\% accuracy on \texttt{presence},
indicating its ability to capture certain cues beyond chance.
However, its performance on the remaining metrics was markedly lower.
The retrieval based method achieved relatively higher overall scores, yet its performance in the \textbf{Fluctuation} category was substantially low. This indicates difficulty in distinguishing different types of fluctuations for TSPulse.
By contrast, all DiffNator configurations substantially outperformed both baselines. 
With respect to merging methods, query attention on difference (attn) surpassed mean pooling (mean) on most metrics. 
The advantage of attention was most evident in the \textbf{Trend} and \textbf{Event} categories, where match accuracy increased by more than 20 percentage points. 
Across encoders, Dilated~TCN+attn and Informer+attn yielded comparable overall match accuracy, with Informer exhibiting a slight overall advantage. 
The \textbf{Event} category remained the most challenging, reflecting the inherent difficulty of localizing short-duration patterns where even small boundary mismatches substantially reduce IoU.

\medskip
\noindent\textbf{Performance under $K_{\max}=2,3,4$.}
Table~\ref{tab:results_k2} reports the performance of the Informer+attn configuration when multiple differences were present. 
As $K_{\max}$ increased from 2 to 4, performance gradually declined across all metrics. 
The reduction was most pronounced in the \textbf{Trend} and \textbf{Event} categories.
This degradation reflects the increased difficulty of localizing short or overlapping patterns when multiple components must be identified simultaneously, and is particularly pronounced for \textbf{Trend}, since this category includes a larger set of component functions, making their identification more challenging.
Despite this decline, performance remained markedly higher than that of the baselines under $K_{\max}{=}1$. 
Furthermore, over- and under-prediction rates increased only moderately with $K_{\max}$, indicating that the framework maintains reasonable robustness even as the number of differences grows.

\section{Conclusion}
We introduced DiffNator, a framework for structured explanations of time-series differences. By classifying differences into two types and representing them in a unified JSON schema, DiffNator provides outputs that are both human-interpretable and machine-actionable. Experiments on the TORI dataset confirmed that our approach outperforms both the VQA baseline and retrieval-based baseline.
\vfill\pagebreak
\section{REFERENCES}

\printbibliography

\end{document}